\documentclass[conference]{IEEEtran}
\IEEEoverridecommandlockouts
\usepackage{cite}
\usepackage{amsmath,amssymb,amsfonts}
\usepackage{algorithmicx}
\usepackage{graphicx}
\usepackage{textcomp}
\usepackage{multirow} 
\usepackage{booktabs}
\usepackage{url}
\usepackage{booktabs}       

\usepackage{amsfonts,amsmath}       
\usepackage{nicefrac}       
\usepackage{microtype}      

\usepackage{multirow}       
\usepackage{graphics}
\usepackage{tabularx}
\usepackage{mathrsfs}
\usepackage{makecell}
\usepackage{caption}
\usepackage{wrapfig}
\usepackage{enumitem}
\usepackage{subcaption}
\usepackage{bbding}

\usepackage{algpseudocode}
\usepackage{amsfonts}       
\usepackage[colorlinks,citecolor=blue,linkcolor=red,bookmarks=false]{hyperref}
\usepackage{color, colortbl}
\usepackage{xcolor}
\definecolor{Gray}{gray}{0.95}
\definecolor{orange}{rgb}{0.9,0.5,0}
\definecolor{LightCyan}{rgb}{0.88,1,1}
\def\BibTeX{{\rm B\kern-.05em{\sc i\kern-.025em b}\kern-.08em
    T\kern-.1667em\lower.7ex\hbox{E}\kern-.125emX}} 
\definecolor{DeepBlack}{RGB}{10, 10, 10}

\begin{document}

\title{MambaMIC: An Efficient Baseline for Microscopic Image Classification with State Space Models}

\author{
	\IEEEauthorblockN{
		Shun Zou\IEEEauthorrefmark{1,5}$^{\ast}$\thanks{$^{\ast}$Equal Contribution, $^{\dagger}$Corresponding author.}, 
		Zhuo Zhang\IEEEauthorrefmark{2}$^{\ast}$, 
        Yi Zou\IEEEauthorrefmark{3},
		Guangwei Gao\IEEEauthorrefmark{4,5}$^{\dagger}$\thanks{This work was supported in part by the Open Fund Project of Provincial Key Laboratory for Computer Information Processing Technology (Soochow University) under Grant KJS2274.}} 
        \IEEEauthorblockA{%
      \begin{tabular}{cc}
        \IEEEauthorrefmark{1}Nanjing Agricultural University & \IEEEauthorrefmark{2}National University of Defense Technology
      \end{tabular}%
    }
    \IEEEauthorblockA{%
  \begin{tabular}{ccc}
    \IEEEauthorrefmark{3}Xiangtan University & \IEEEauthorrefmark{4}Nanjing University of Posts and Telecommunications & \IEEEauthorrefmark{5}Soochow University
  \end{tabular}%
}

	\IEEEauthorblockA{
 zs@stu.njau.edu.cn, zhangzhuo@nudt.edu.cn, 202205570112@smail.xtu.edu.cn, csgwgao@njupt.edu.cn}
}

\DeclareRobustCommand*{\IEEEauthorrefmark}[1]{%
    \raisebox{0pt}[0pt][0pt]{\textsuperscript{\footnotesize\ensuremath{#1}}}}

\maketitle

\begin{abstract}
In recent years, CNN and Transformer-based methods have made significant progress in Microscopic Image Classification (MIC). However, existing approaches still face the dilemma between global modeling and efficient computation. While the Selective State Space Model (SSM) can simulate long-range dependencies with linear complexity, it still encounters challenges in MIC, such as local pixel forgetting, channel redundancy, and lack of local perception. To address these issues, we propose a simple yet efficient vision backbone for MIC tasks, named MambaMIC. Specifically, we introduce a Local-Global dual-branch aggregation module: the MambaMIC Block, designed to effectively capture and fuse local connectivity and global dependencies. In the local branch, we use local convolutions to capture pixel similarity, mitigating local pixel forgetting and enhancing perception. In the global branch, SSM extracts global dependencies, while Locally Aware Enhanced Filter reduces channel redundancy and local pixel forgetting. Additionally, we design a Feature Modulation Interaction Aggregation Module for deep feature interaction and key feature re-localization. Extensive benchmarking shows that MambaMIC achieves state-of-the-art performance across five datasets. code is available at \href{https://zs1314.github.io/MambaMIC}{https://zs1314.github.io/MambaMIC}

\end{abstract}


\section{Introduction}
Microscopic imaging technology plays a crucial role in the medical field and is an indispensable tool in modern medical research and clinical diagnosis \cite{ying2006modern}. By classifying microscopic images, medical researchers can observe the structural and dynamic changes at the tissue, cellular, and molecular levels, leading to a deeper understanding of disease mechanisms \cite{merchant2022microscope}. 

In recent years, inspired by the success of deep learning in various vision tasks, many studies have developed different network architectures based on Convolutional Neural Networks (CNNs) and applied them to MIC \cite{LIU2021104523,kumar2023advances,nguyen2018deep}. Although convolution operations can effectively model local connectivity, their inherent characteristics, such as limited local receptive fields, hinder the extraction of long-range dependencies, often resulting in insufficient semantic context extraction and incomplete feature representation. Fortunately, inspired by the Transformer in natural language processing and advanced vision tasks \cite{vaswani2017attention}, 
\begin{figure}[ht]
	\centering
	\includegraphics[width=0.5\textwidth]{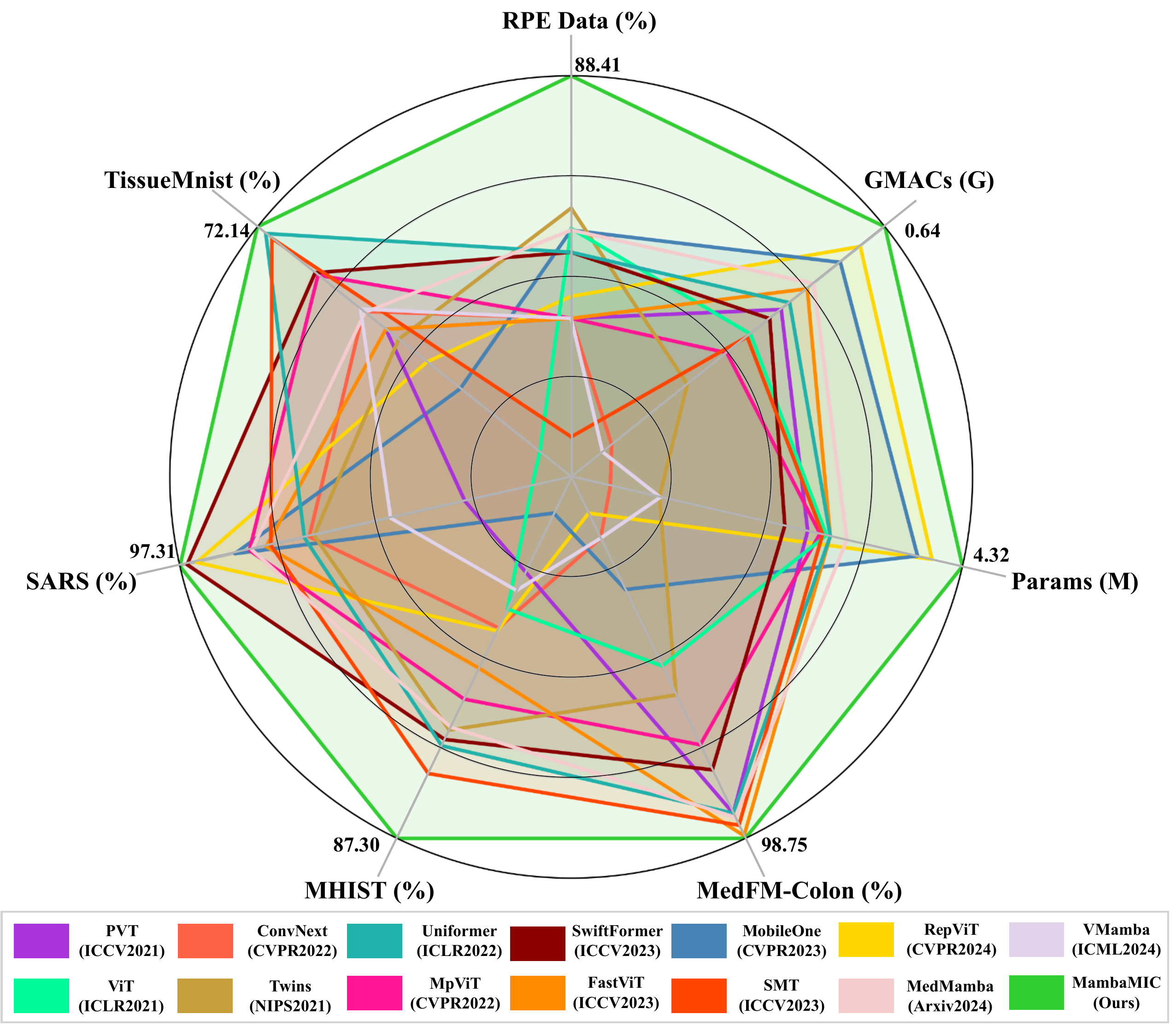} 
	\vspace{-5mm}
	\caption{A seven-dimensional radar map of the Overall Accuracy of RPE Data \cite{nanni2016texture}, TissueMnist \cite{medmnistv2}, SARS \cite{Yu2023}, MHIST \cite{Wei2021}, MedFM-Colon \cite{wang2023real}, along with Params and GMACs.}
	\label{compare}
	\vspace{-7mm}
\end{figure}
Transformer-based architectures have been developed for MIC tasks \cite{LIU2021104523,sengar2022vision,abou2023white}. Thanks to the self-attention mechanism, they can effectively model global dependencies, alleviating the limitations of CNN models. However, Transformer-based methods still face a significant challenge: they exhibit high quadratic complexity when modeling long sequences, leading to substantial computational overhead. This disregards the computational constraints in real-world medical environments and fails to meet the need for low-parameter, low-computation models in mobile MIC \cite{masud2020leveraging}. While some studies have adopted efficient attention techniques, such as mobile window attention \cite{tu2022maxvit,huang2022lightvit}, these approaches still fail to fully exploit the information within each patch, often sacrificing the global receptive field and not fundamentally resolving the trade-off between global dependency modeling and efficient computation.

Recently, State Space Models (SSM) \cite{kalman1960new,gupta2022diagonal} have attracted significant attention from researchers. Building upon classical SSM research, modern SSMs like Mamba \cite{gu2023mamba} not only establish long-range dependencies but also exhibit linear complexity with respect to input size, making Mamba a strong competitor to CNNs and Transformers in lightweight MIC tasks. However, Mamba still faces three major challenges when applied to MIC: (1) Since Mamba processes the flattened 1D image sequence in a recursive manner, it may cause adjacent pixels in the 2D space to be far apart in the flattened sequence, leading to local pixel forgetting. In MIC tasks, adjacent pixels often have strong relationships, and the loss of this relationship results in the loss of semantic information in multiple local blocks, creating a cumulative effect and leading to a disastrous pixel forgetting phenomenon; (2) Due to the need to memorize long-range dependencies in Visual State Space Models (VSSM), the number of hidden states in the state space equations becomes very large, which not only leads to channel information redundancy and increased computational burden, but also generates a significant amount of irrelevant interference, thereby hindering the representation learning of critical channel information; (3) Unlike other vision tasks with clearly defined target features, MIC requires not only the capture of global context but also a focus on local fine-grained features.

\begin{figure}[ht]
	\centering
	\includegraphics[width=0.5\textwidth]{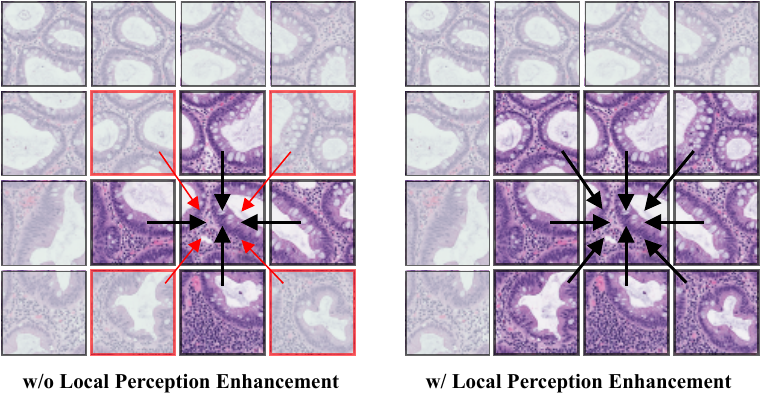} 
	\vspace{-6mm}
	\caption{In Mamba's one-dimensional recursive image processing, local pixels (highlighted in \textcolor{red}{red}) are easily forgotten in the flattened sequence. However, enhancing local perception effectively captures pixel relationships.}
	\label{local}
	\vspace{-7mm}
\end{figure}

To address the above issues, we introduce MambaMIC, a simple yet highly effective baseline model. Its core idea is to fully leverage the local feature extraction advantages of CNNs and the global modeling strengths of Mamba, while maintaining linear complexity and a low computational burden. Specifically, the core component of MambaMIC is the MambaMIC Block, which adopts a Local-Global dual-branch architecture designed to effectively extract and aggregate both local invariant features and long-range dependency characteristics. 
In the Local branch, we use local convolutional designs that, on the one hand, capture fine-grained local features, providing local connectivity and enhancing local perception ability, and on the other hand, alleviate the local pixel forgetting problem faced by vanilla Mamba \cite{gu2023mamba} when dealing with 2D images (see Fig. \ref{local}). The Global branch, composed of multiple parallel Residual Efficient Vision State Space Modules (REVSSM), mitigates information blocking caused by the exponential increase in hidden state numbers as the number of channels grows, thanks to the parallel mechanism. Moreover, in the REVSSM, we introduce the Locally Aware Enhanced Filter (LAEF), which employs a sophisticated channel selection and pruning mechanism to enhance the local perception capability of VSSM, promote context expert information interaction and flow, and reduce channel redundancy caused by excessive hidden states, allowing the most valuable information to circulate globally. Simultaneously, LAEF and the local convolutions of the Local branch form a complementary flow, enhancing local pixel blocks through both parallel and serial paradigms.
Additionally, we observe a non-negligible feature gap between the Local and Global branches. Simple addition or concatenation inevitably leads to the loss of valuable information, limiting performance improvements. Therefore, to further promote feature fusion and information interaction within the paradigm, we propose the Feature Modulation Interaction Aggregation Module (FMIAM). FMIAM achieves deep fusion and interaction by adaptively weighting the corresponding branches, and we also incorporate a simplified channel attention mechanism to recalibrate and localize channel features, filtering out irrelevant features and enhancing the representation of key features. 
Finally, comprehensive experiments demonstrate that MambaMIC achieves the optimal performance-parameter trade-off, making it a true "jack-of-all-trades," as shown in Fig. \ref{compare}.
In summary, our main contributions can be summarized as follows:  
\vspace{-1mm}
\begin{itemize}
    \item We are the first to apply State Space Models (SSM) to MIC through extensive experiments, leading to the proposal of MambaMIC, a simple yet effective alternative to CNN- and Transformer-based methods.  
    \item We introduce a simple and efficient dual-branch architecture, the MambaMIC Block, consisting of a local branch and a global branch. 
    Specifically, we develop the Residual Efficient Vision State Space Module as the core of the global branch and enhance local perception using the Locally Aware Enhanced Filter, promoting the interaction and flow of contextual channel information while reducing channel redundancy caused by excessive hidden states. Additionally, we introduce the Feature Modulation Interaction Aggregation Module to effectively bridge the semantic gap between different types of features and better aggregate diverse information.  
    \item Extensive experiments on five datasets demonstrate that our MambaMIC outperforms other state-of-the-art methods, providing a new benchmark and reference for MIC.
\end{itemize}
\begin{figure*}[ht]
	\centering
	\includegraphics[width=\textwidth]{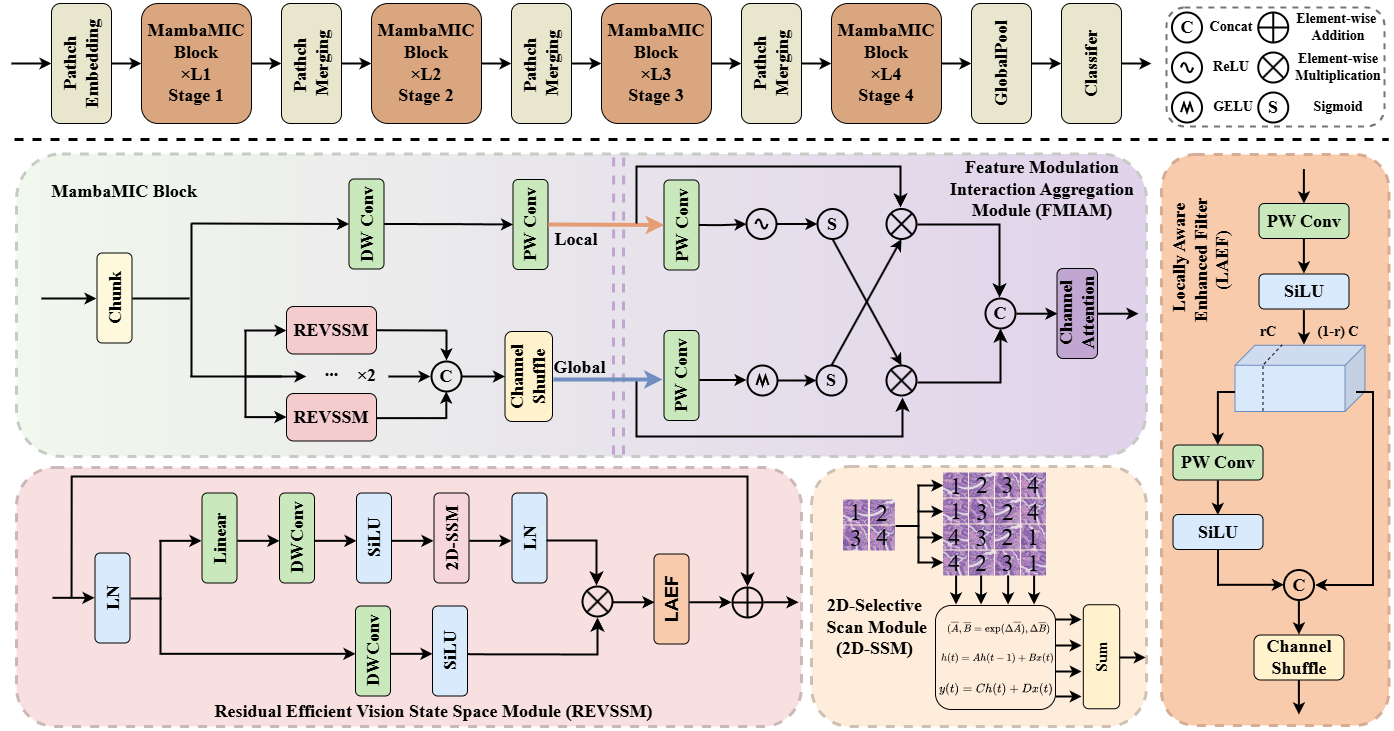} 
	\vspace{-6mm}
	\caption{The overall architecture of the proposed MambaMIC.}
	\label{overall}
	\vspace{-7mm}
\end{figure*}
\section{Method}
\subsection{Overall Pipeline}
Fig. \ref{overall} illustrates the overall architecture of MambaMIC. Consistent with previous general visual backbones \cite{Shaker_2023_ICCV,lin2023scaleaware,liu2024vmamba}, MambaMIC is divided into four stages, each consisting of several stacked MambaMIC Blocks. Additionally, each stage is preceded by an Embedding or Merging layer for spatial downsampling and channel expansion. A global average pooling layer is applied to the final output, which is then fed into a linear classification head. 
The MambaMIC Block is the core component of MambaMIC. Specifically, as shown in Fig. \ref{overall}, it adopts a Local-Global dual-branch interactive aggregation architecture. In the global branch, we introduce four parallel Residual Efficient Vision State Space Modules (REVSSM). The parallel mechanism alleviates the computational burden caused by the exponential increase in the number of states with the number of channels in the visual state space, while also promoting the interaction of contextual channel information, compensating for the lack of global relationship modeling at the channel level in VSSM.  
Additionally, we observed that since SSM processes the flattened feature map into a 1D sequence of tokens, the proximity of adjacent pixels in the sequence is highly influenced by the flattening strategy. For instance, when using the four-direction unfolding strategy from \cite{liu2024vmamba}, only four nearest adjacent pixels are available for each anchor, meaning some spatially neighboring pixels in the 2D feature map are far apart in the 1D token sequence \cite{guo2025mambair}. This long-distance separation leads to local pixel forgetting, where the relationships between pixels gradually diminish (see Fig. \ref{local}). To address the missing local information in Mamba, we introduce additional local convolutions in the local branch to help restore pixel neighborhood similarity. 
Furthermore, we develop the Feature Modulation Interaction Aggregation Module (FMIAM) to reduce the knowledge gap between the local and global branches, fully fuse internal paradigm features, and implement channel recalibration via a simplified channel attention mechanism. Mathematically, the entire process of the MambaMIC Block can be expressed as follows:  
\vspace{-2mm}
\begin{equation}
    F',F''=Chunk(F),
    \vspace{-2mm}
\end{equation}
\begin{equation}
    F_{L}=PW(DW(F')),
    \vspace{-2mm}
\end{equation}
\begin{equation}
    F_{1},F_{2},F_{3},F_{4}=Split(F''),
    \vspace{-2mm}
\end{equation}
\begin{equation}
    \hat{F_{i}} =REVSSM(F_{i}),i\in [1,2,3,4],
    \vspace{-2mm}
\end{equation}
\begin{equation}
    F_{G}=Shuffle([\hat{F_{1}},\hat{F_{2}},\hat{F_{3}},\hat{F_{4}} ]),
    \vspace{-2mm}
\end{equation}
\begin{equation}
    \hat{F}=FMIAM(F_{L},F_{G}).
    \vspace{-2mm}
\end{equation}
where Chunk(·) represents channel splitting, DW(·) denotes Depthwise Separable Convolution, PW(·) refers to Point-Wise Convolution, Shuffle(·) indicates Channel Shuffle, [,] represents channel concatenation, REVSSM(·) stands for Residual Efficient Vision State Space Module, and FMIAM(·) denotes Feature Modulation Interaction Aggregation Module. 
\subsection{Residual Efficient Vision State Space Module}
\label{REVSSM}
Fig. \ref{overall} illustrates the Residual Efficient Vision State Space Module (REVSSM), which utilizes state space equations to capture long-range dependencies. Specifically, given the input feature \( X \in \mathbb{R}^{H \times W \times C} \), after layer normalization, the features pass through two parallel branches.
In the first branch, the feature channels undergo Depthwise Separable Convolution (DWConv) and are expanded to \( \lambda C \) (where \( \lambda \) is the channel expansion factor), followed by processing with the SiLU activation function. 
In the second branch, the feature channels are first expanded to \( \lambda C \) via a linear layer, then processed with DWConv, SiLU activation, a 2D-selective scanning module (2D-SSM), and a LayerNorm layer. The features from both branches are then roughly aggregated through element-wise multiplication. 
\begin{table*}[htp]
  \centering
  \caption{Comparison with state-of-the-art methods on five public datasets \cite{nanni2016texture,Yu2023,medmnistv2,Wei2021,wang2023real}. The best results are highlighted in \textbf{bold fonts}. “ ↑ ”and “ ↓ ” indicate that larger or smaller is better.}
  \vspace{-2mm}
    \begin{tabular}{l l c p{0.8cm} p{0.8cm} >{\centering\arraybackslash}p{0.6cm} >{\centering\arraybackslash}p{0.6cm} >{\centering\arraybackslash}p{0.6cm} >{\centering\arraybackslash}p{0.6cm} >{\centering\arraybackslash}p{0.6cm} >{\centering\arraybackslash}p{0.6cm} >{\centering\arraybackslash}p{0.6cm} >{\centering\arraybackslash}p{0.6cm} >{\centering\arraybackslash}p{0.6cm} >{\centering\arraybackslash}p{0.5cm}}
    \toprule
     & \multirow{2}[2]{*}{\centering Method} & \multirow{2}[2]{*}{\centering Year} & \multirow{2}[2]{*}{GMACs$\downarrow$}& \multirow{2}[2]{*}{Params$\downarrow$} & \multicolumn{2}{c}{RPE Data \cite{nanni2016texture}} & \multicolumn{2}{c}{MHIST \cite{Wei2021}} & \multicolumn{2}{c}{SARS \cite{Yu2023}} & \multicolumn{2}{c}{TissueMnist \cite{medmnistv2}} & \multicolumn{2}{c}{FM-Colon \cite{wang2023real}} \\
\cmidrule{6-15} & & & & & \multicolumn{1}{>{\centering\arraybackslash}p{0.5cm}}{OA$\uparrow$} & \multicolumn{1}{>{\centering\arraybackslash}p{0.6cm}}{AUC$\uparrow$} & \multicolumn{1}{>{\centering\arraybackslash}p{0.5cm}}{OA$\uparrow$} & \multicolumn{1}{>{\centering\arraybackslash}p{0.6cm}}{AUC$\uparrow$} & \multicolumn{1}{>{\centering\arraybackslash}p{0.5cm}}{OA$\uparrow$} & \multicolumn{1}{>{\centering\arraybackslash}p{0.6cm}}{AUC$\uparrow$} & \multicolumn{1}{>{\centering\arraybackslash}p{0.5cm}}{OA$\uparrow$} & \multicolumn{1}{>{\centering\arraybackslash}p{0.6cm}}{AUC$\uparrow$} & \multicolumn{1}{>{\centering\arraybackslash}p{0.5cm}}{OA$\uparrow$} & \multicolumn{1}{>{\centering\arraybackslash}p{0.4cm}}{AUC$\uparrow$} \\

    \midrule
    \multirow{9}{*}{\rotatebox{90}{CNNs}}& ConvNext-tiny\cite{liu2022convnet} & CVPR2022 & 4.49 & 28.69 & 86.06& 97.96& 77.34& 84.24& 96.88& 99.46& 66.47& 92.07& 94.56& 98.71\\
    & ConvNext-small \cite{liu2022convnet} & CVPR2022 & 8.73 & 50.22 & 85.44& 98.02& 76.55& 84.12& 96.77& 99.42& 69.22& 93.21& 87.36& 95.05\\
    & RepViT-m1\_0\cite{wang2024repvit} & CVPR2024 & 1.13 & 6.85 & 87.60& 98.15& 69.10& 71.32& 96.62& 99.42& 67.54& 92.45& 97.45& 99.63\\
    & RepViT-m1\_1 \cite{wang2024repvit} & CVPR2024 & 1.37 & 8.29 & 85.71& 98.06& 76.70& 80.90& 97.24& 99.56& 67.47& 92.41& 86.41& 93.60\\
    & MobileOne-s0\cite{mobileone2022} & CVPR2023 & 1.10& 5.29 & 86.79& 98.28& 67.83& 60.04& 96.03& 99.17& 63.20& 90.34& 89.26& 95.97\\
    & MobileOne-s2 \cite{mobileone2022} & CVPR2023 & 1.35 & 7.88 & 85.71& 98.16& 70.68& 73.07& 96.61& 99.33& 62.66& 90.11& 87.76& 94.62\\
    & MobileOne-s3 \cite{mobileone2022} & CVPR2023 & 1.96 & 10.17 & 86.52& 98.24& 70.69& 73.44& 97.08& 99.47& 66.55& 92.05& 89.31& 95.78\\
    \midrule
    \multirow{9}{*}{\rotatebox{90}{ViTs}}& PVT-smal\cite{wang2021pvtv2} & ICCV2021 & 3.71 & 24.49 & 85.44& 98.13& 73.69& 80.11& 96.12& 99.16& 68.60& 92.88& 97.80& 99.81\\
    & PVT-medium \cite{wang2021pvtv2} & ICCV2021 & 6.49 & 44.21 & 87.33& 98.19& 80.82& 87.72& 96.61& 99.31& 69.17& 93.14& 97.25& 99.57\\
    & MpViT-tiny\cite{lee2022mpvit} & CVPR2022 & 1.84 & 5.84 & 87.87& 98.11& 81.30& 87.37& 96.93& 99.43& 70.63& 93.79& 98.05& 99.83\\
    & MpViT-small \cite{lee2022mpvit} & CVPR2022 & 5.32 & 22.89 & 85.44& 97.95& 80.19& 86.57& 97.02& 99.52& 70.47& 93.73& 95.20& 98.51\\
    & Twins-small\cite{chu2021Twins} & NIPS2021 & 3.71 & 24.11 & 88.14& 98.24& 78.76& 86.15& 96.61& 99.40& 67.80& 82.55& 97.45& 99.65\\
    & Twins-base \cite{chu2021Twins} & NIPS2021 & 6.49 & 43.83 & 86.79& 98.40& 81.77& 88.93& 96.75& 99.43& 68.25& 92.82& 93.31& 98.29\\
    & ViT-tiny\cite{dosovitskiy2020image} & ICLR2021 & 1.26 & 5.71 & 86.52& 97.50& 74.33& 79.33& 96.01& 99.18& 58.50& 87.80& 96.15& 99.38\\
    & ViT-small \cite{dosovitskiy2020image} & ICLR2021 & 4.62 & 22.04 & 86.54& 97.76& 75.59& 80.94& 95.84& 99.20& 64.36& 90.94& 92.21& 97.63\\
    & ViT-base \cite{dosovitskiy2020image} & ICLR2021 & 17.6 & 86.54 & 86.24& 97.25& 75.44& 82.59& 96.43& 99.37& 65.86& 91.73& 94.61& 94.60\\
    \midrule
    \multirow{9}{*}{\rotatebox{90}{Hybrid-CNN-ViT}}& FastViT-sa24\cite{vasufastvit2023} & ICCV2023 & 2.94 & 21.55 & 85.44& 98.07& 78.61& 84.04& 96.95& 99.43& 68.60& 92.92& 98.65& 99.89\\
    & FastViT-ma36 \cite{vasufastvit2023} & ICCV2023 & 6.07 & 44.07 & 88.14& 98.08& 81.93& 85.55& 96.95& 99.45& 68.79& 93.00& 94.31& 98.43\\
    & SwiftFormer-S\cite{Shaker_2023_ICCV} & ICCV2023 & 1.01 & 5.64 & 85.44& 98.04& 81.62& 90.16& 96.88& 99.45& 69.57& 93.40& 94.91& 98.77\\
    & SwiftFormer-L1 \cite{Shaker_2023_ICCV} & ICCV2023 & 1.62 & 11.29 & 86.52& 98.02& 81.93& 88.47& 97.21& 99.54& 70.59& 93.66&  95.45&  98.92\\
    & SwiftFormer-L3 \cite{Shaker_2023_ICCV} & ICCV2023 & 4.05 & 27.47 & 86.25& 98.04& 82.25& 89.95& 97.28& 99.53& 70.55& 93.58& 96.15& 99.27\\
    & Uniformer-small\cite{li2022uniformer} & ICLR2022 & 3.46 & 21.55 & 86.25& 88.13& 82.57& 89.96& 96.79& 99.45& 71.91& 94.27&  97.80&  99.71\\
    & Uniformer-base \cite{li2022uniformer} & ICLR2022 & 7.81 & 49.78 & 88.14& 83.41& 81.77& 89.08& 97.22& 99.49& 72.06& 94.08& 95.80& 99.15\\
    & SMT-s\cite{lin2023scaleaware} & ICCV2023 & 4.72 & 22.55 & 83.99 & 83.73 & 83.99& 89.73& 96.93& 99.46& 71.74& 94.33&  98.25&  99.84\\
    & SMT-b \cite{lin2023scaleaware} & ICCV2023 & 7.81 & 32.04 & 87.87& 98.10& 86.05& 91.04& 96.90& 99.45& 69.27& 93.36& 97.85& 99.73\\
    \midrule
    \multirow{4}{*}{\rotatebox{90}{Mambas}}& Medmamba-s\cite{yue2024medmamba} & Arxiv2024 & 2.75 & 19.39 & 86.52& 98.17& 81.62& 87.04& 97.01& 99.23& 69.18& 93.12&  97.95&  99.85\\
    & Medmamba-b \cite{yue2024medmamba} & Arxiv2024 & 6.16 & 40.88 & 86.79& 98.02& 77.97& 85.20& 97.30& 99.51& 69.11& 93.18& 95.70& 98.94\\
    & VMamba-t\cite{liu2024vmamba} & ICML2024 & 4.4 & 22.1 & 85.71& 97.80& 77.34& 83.32& 95.92& 99.28& 69.23& 93.13& 92.66& 97.84\\
    & VMamba-s \cite{liu2024vmamba} & ICML2024 & 9.0 & 43.7 & 85.44& 97.79& 74.64& 81.27& 96.43& 99.37& 69.30& 93.20& 87.36& 94.77\\
     \midrule
    \multirow{3}{*}{\rotatebox{90}{Ours}}& MambaMIC-t & -& \pmb{0.64} & \pmb{4.32} & \pmb{88.41}& \pmb{98.29}& \underline{87.30}& \underline{93.49}& \pmb{97.31}& \pmb{99.60}& \underline{72.14}& \underline{94.31}&  \pmb{98.75}&  \pmb{99.90}\\
    & MambaMIC-s & -& 0.76 & 4.97 & 87.06& 98.24& 86.05& 92.31& 97.13& 99.49& \pmb{73.01}& \pmb{94.40}& 98.00& 99.78\\
    & MambaMIC-b & -& 1.59 & 8.39 & 87.33& 98.17& \pmb{87.80}& \pmb{94.17}& 97.14& 99.48& 70.53& 93.50& 96.85& 99.64\\
    \bottomrule
    \end{tabular}
  \label{table1}
\vspace{-6mm}
\end{table*} 
Finally, the rough-aggregated features pass through the Locally Aware Enhanced Filter (LAEF) to enhance local information perception and capture pixel neighborhood similarity, while mitigating channel information redundancy. Residual connections are introduced, resulting in the output \( \hat{X} \). The process is formally expressed as follows:
\vspace{-2mm}
\begin{equation}
    X_{1}=LN(2D\text{-} SSM(SiLU(DW(Linear(X))))),
    \vspace{-2mm}
\end{equation}
\begin{equation}
    X_{2}=SiLU(DW(LN(X))),
    \vspace{-2mm}
\end{equation}
\begin{equation}
    \hat{X}=X+LAEF(X_{1}\odot X_{2}),
    \vspace{-2mm}
\end{equation}
where Linear(·) represents processing using a linear layer, LN(·) denotes the layer normalization process, SiLU(·) is the SiLU activation function, 2D-SSM(·) denotes the 2D Selective Scanning Module, \(\odot\) represents element-wise multiplication, and LAEF(·) denotes the Locally Aware Enhanced Filter.
\vspace{-2mm}
{\flushleft\textbf{Locally Aware Enhanced Filter}.}
Due to the flattening characteristic of vanilla Mamba’s feature operations \cite{liu2024vmamba}, it leads to pixel forgetting within local regions when handling 2D images. This induces an accumulation effect of pixel information loss, causing incoherence in key semantic information and severely impacting the correct understanding of the image. To address this, we previously introduced a local branch that uses local convolutions to enhance pixel similarity. However, we believe this improvement is still insufficient to fully solve the problem. 
Moreover, we found that Mamba requires the memory of long-sequence dependencies, which causes the number of hidden states in the state-space equations to accumulate significantly. This results in redundant information, where irrelevant information not only adds extra computational burden but also interferes with the representation learning of key features, preventing effective flow of crucial expert information. To address this, we developed the Locally Aware Enhanced Filter (LAEF) in the Visual State Space Model (VSSM), which employs a graceful channel routing and local enhancement mechanism to further resolve these challenges. Fig. \ref{overall} shows the architecture of LAEF.
Specifically, given the input feature \( X \in \mathbb{R}^{H \times W \times C} \), the input is first embedded into a lower-dimensional space through Point-Wise Convolution (PWConv), followed by a SiLU activation function to obtain \( X' \). Then, \( X' \) is split along the channel dimension into two groups: one for local information perception and the other for retained information. The number of channels in the local information perception group is set to \( rC \), while the number of channels in the retained information group is set to \( (1-r)C \), where \( r \) is the partial channel ratio (the specific setting of \( r \) will be elaborated in the experiments section). In the local information perception group, the divided features are sequentially processed by PWConv and SiLU activation functions to enhance local information. Finally, the locally enhanced features are concatenated with the retained features, followed by channel shuffle. Formally, the above process is defined as follows:
\begin{equation}
    X'=SiLU(PW(X)),
    \vspace{-2mm}
\end{equation}
\begin{equation}
    X_{L} = X' \mid  _{:, :, [1, rC]}, \quad X_{R} = X'\mid _{:, :, [rC+1, C]},
    \vspace{-2mm}
\end{equation}
\begin{equation}
    \hat{X}=Shuffle(SiLU(PW(X_{L})),X_{R}).
    \vspace{-2mm}
\end{equation}

\subsection{Feature Modulation Interaction Aggregation Module}
\label{FMIAM}
Although we capture sufficient representation information through the local and global branches, effectively integrating the information from both branches becomes a critical challenge. In fact, an intuitive observation is that there exists an uncertain knowledge gap between the convolution-based CNN local features and the SSM-based global features. Therefore, simply adding or concatenating these features does not fully exploit their potential.
\vspace{-2mm}
\begin{figure}[ht]
	\centering
	\includegraphics[width=0.4\textwidth]{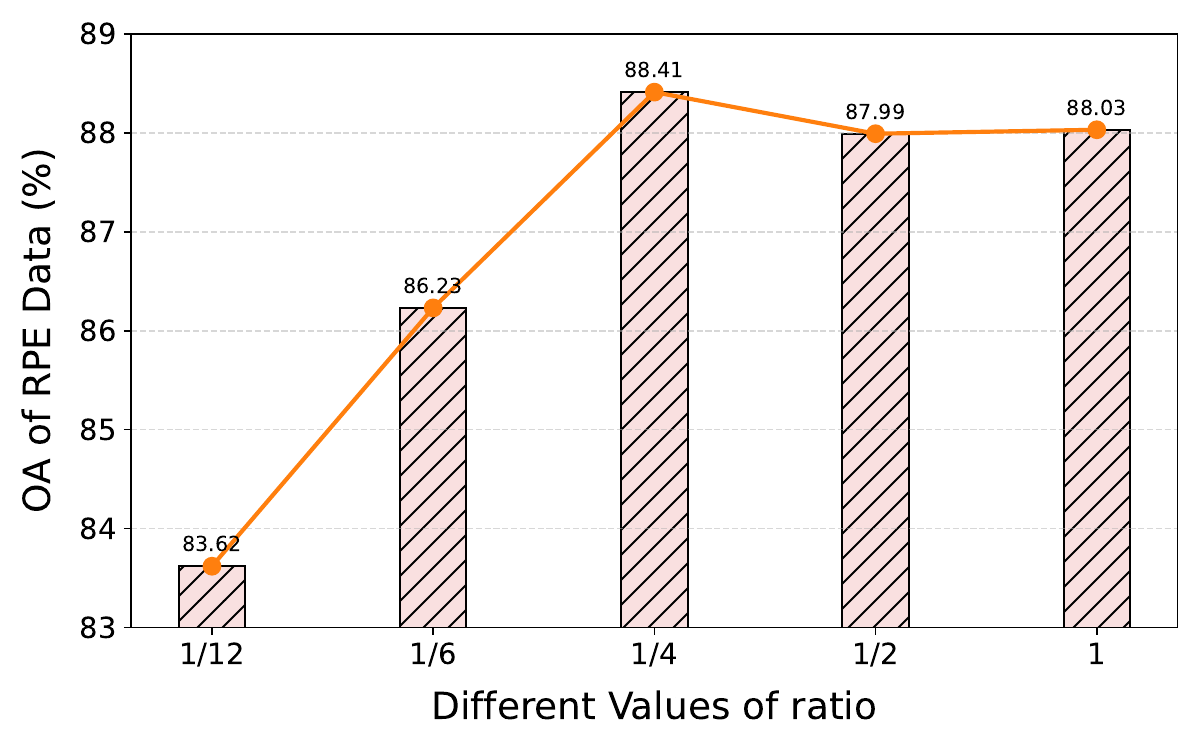} 
	\vspace{-3mm}
	\caption{Ablation analysis of different values of partial ratio.}
	\label{rate}
	\vspace{-4mm}
\end{figure}
\begin{table}[htp]
    \centering
    \setlength{\abovecaptionskip}{-0cm}
    \caption{Quantitative comparisons with different combinations of the LAEF and FMIAM.}
    \begin{tabular}{>{\centering\arraybackslash}p{0.5cm} >{\centering\arraybackslash}p{0.8cm} | >{\centering\arraybackslash}p{0.7cm} >{\centering\arraybackslash}p{0.5cm} >{\centering\arraybackslash}p{0.5cm} >{\centering\arraybackslash}p{0.5cm} >{\centering\arraybackslash}p{0.5cm} >{\centering\arraybackslash}p{0.5cm} >{\centering\arraybackslash}p{0.5cm}}
        \toprule[1.0pt]
        \multirow{2}[2]{*}{LAEF} & \multirow{2}[2]{*}{FMIAM} &\multirow{2}[2]{*}{Params} & \multicolumn{3}{c}{RPE Data \cite{nanni2016texture}} & \multicolumn{3}{c}{MHIST \cite{Wei2021}} \\
        \cmidrule(lr){4-6} \cmidrule(lr){7-9}
        &  & & OA & Pre & AUC & OA & Pre & AUC \\
        \midrule[0.5pt]
        \XSolid     & \XSolid     & 4.86  &       87.14&       88.21&       97.32&       84.79&       84.56&  91.10\\
        \Checkmark  & \XSolid     & 4.30  & 87.87 & 89.24 & 98.28 &       86.21&       85.99&  89.51\\
        \XSolid     & \Checkmark  & 4.89  & 87.60 & 89.28 & 98.15 &       85.74&       85.92&  91.83\\
        \Checkmark  & \Checkmark  & 4.32  & 88.41 & 89.68& 98.29 &       87.30&       87.67&  93.49\\
        \bottomrule
    \end{tabular}
    \label{ablation1}
\vspace{-6mm}
\end{table}
To address this, we introduce the FMIAM, which operates between the two branches to reweight the features, enabling deep interaction and aggregation of the information. Specifically, we first compute the globally weighted features and locally weighted features, then modulate and reweight the features from both branches, and finally introduce a simplified channel attention mechanism \cite{wang2020eca} to reduce interference from irrelevant information and achieve key feature localization. The above process is expressed as follows:
\begin{equation}
    W_{L}=Sigmoid(\varphi (PW(F_{Local}))),
    \vspace{-2mm}
\end{equation}
\begin{equation}
    W_{G}=Sigmoid(\varrho (PW(F_{Global}))),
    \vspace{-3mm}
\end{equation}
\begin{equation}
    W_{fusion}=CA([W_{G}\odot F_{Local},W_{L}\odot F_{Global}]).
    \vspace{-2mm}
\end{equation}
where \( \text{Sigmoid}(\cdot) \) denotes the Sigmoid activation function, \( \varphi(\cdot) \) represents the ReLU activation function, \( \varrho(\cdot) \) represents the GELU activation function, and \( \text{CA}(\cdot) \) denotes the channel attention mechanism.
\section{Experiments}

\subsection{Experimental settings}
{\flushleft\textbf{Datasets}.}
To validate the model's performance, we selected five public medical image datasets:  the Retinal Pigment Epithelium Cell dataset (RPE Data) \cite{nanni2016texture}, minimalist histopathology image analysis dataset (MHIST) \cite{Wei2021}, the Malaria Cell Image dataset (SARS) \cite{Yu2023}, TissueMNIST \cite{medmnistv2}, and MedFM-Colon \cite{wang2023real}. The RPE dataset comprises 1,862 images of retinal pigment epithelium cells, classified into four categories. The SARS dataset includes 27,558 images of malaria cells, divided into two categories. MHIST contains 3,152 images of colorectal polyps for binary classification. The MedMF-Colon dataset classifies tumor tissues in colonoscopy pathology slides, consisting of 10,009 pathological tissue patches from 396 colon cancer patients. TissueMNIST features 236,386 images of human kidney cortex cells, categorized into eight groups. All datasets were split into training, validation, and test sets in a 6:2:2 ratio.
\begin{table}[htp]
    \centering
    \setlength{\abovecaptionskip}{-0cm}
    \caption{Effect of the parallel VSSM mechanism.}
    \begin{tabular}{ >{\centering\arraybackslash}p{1.0cm} | >{\centering\arraybackslash}p{0.7cm} >{\centering\arraybackslash}p{0.5cm} >{\centering\arraybackslash}p{0.5cm} >{\centering\arraybackslash}p{0.5cm} >{\centering\arraybackslash}p{0.5cm} >{\centering\arraybackslash}p{0.5cm} >{\centering\arraybackslash}p{0.5cm}}
        \toprule[1.0pt]
        \multirow{2}[2]{*}{Parallel}  &\multirow{2}[2]{*}{Params} & \multicolumn{3}{c}{RPE Data \cite{nanni2016texture}} & \multicolumn{3}{c}{MHIST \cite{Wei2021}} \\
        \cmidrule(lr){3-5} \cmidrule(lr){6-8}
        & & OA & Pre & AUC & OA & Pre & AUC \\
        \midrule[0.5pt]
        \XSolid        &   8.83&       87.60&      88.71&       98.28&       86.05&       85.83&  91.30\\
        \Checkmark    & 4.32& 88.41& 89.68& 98.29&       87.30&       87.67&  93.49\\
        \bottomrule
    \end{tabular}
    \label{ablation2}
\vspace{-3mm}
\end{table}
\begin{table}[htp]
    \centering
    \setlength{\abovecaptionskip}{-0cm}
    \caption{Ablation studies on alternatives to the LAEF.}
    \begin{tabular}{>{\centering\arraybackslash}p{1.9cm} | >{\centering\arraybackslash}p{0.6cm} >{\centering\arraybackslash}p{0.5cm} >{\centering\arraybackslash}p{0.5cm} >{\centering\arraybackslash}p{0.5cm} >{\centering\arraybackslash}p{0.5cm} >{\centering\arraybackslash}p{0.5cm} >{\centering\arraybackslash}p{0.5cm}}
        \toprule[1.0pt]
        \multirow{2}[2]{*}{Method}  &\multirow{2}[2]{*}{Params} & \multicolumn{3}{c}{RPE Data \cite{nanni2016texture}} & \multicolumn{3}{c}{MHIST \cite{Wei2021}} \\
        \cmidrule(lr){3-5} \cmidrule(lr){6-8}
        &  & OA & Pre & AUC & OA & Pre & AUC \\
        \midrule[0.5pt]
        Linear (baseline) & 4.89  &  87.60&  89.28&       98.15& 85.74& 85.92&  91.83\\
        3\( \times \)3 Conv   & 4.94  & 86.14 & 88.25 & 97.99 &       84.21&   84.02&  89.88\\
        ConvGLU \cite{shi2023transnext}     & 5.35  & 87.98 & 89.88 & 97.52 &       86.34&       86.22&  90.11\\
        DFN \cite{li2021localvit} & 5.27  & 87.25 & 88.69& 97.68 &       85.02&       84.97&  90.08\\
        LAEF (ours) & 4.32  & 88.41 & 89.68& 98.29 &       87.30&       87.67&  93.49\\
        \bottomrule
    \end{tabular}
    \label{table:LAEF}
\vspace{-6mm}
\end{table}

\vspace{-6mm}
{\flushleft\textbf{Training Details}.}
We implemented our MambaMIC with PyTorch 2.0.0 and trained it on an NVIDIA RTX 3090, processing 200 epochs with a batch size of 16. We employed the Adam optimizer with an initial learning rate of 0.0001, a weight decay of 1e-4, and Cross-Entropy Loss to optimize the model parameters. Additionally, we incorporated a cosine annealing learning rate decay strategy and an early stopping strategy with a 10-epoch warm-up period during training.

\subsection{Comparison with SOTA Models}
To validate the effectiveness of MambaMIC, we compared it with state-of-the-art methods, including CNN-based approaches (ConvNext \cite{liu2022convnet}, RepViT \cite{wang2024repvit}, MobileOne \cite{mobileone2022}), Transformer-based approaches (PVT \cite{wang2021pvtv2}, MpViT \cite{lee2022mpvit}, Twins \cite{chu2021Twins}, ViT \cite{dosovitskiy2020image}), hybrid CNN-Transformer methods (FastViT \cite{vasufastvit2023}, SwiftFormer \cite{Shaker_2023_ICCV}, Uniformer \cite{li2022uniformer}, SMT \cite{lin2023scaleaware}), and Mamba-based methods (MedMamba \cite{yue2024medmamba}, VMamba \cite{liu2024vmamba}). As shown in Table \ref{table1} and Fig. \ref{compare}, MambaMIC achieves the best results in terms of model parameters, FLOPs, OA, and AUC across five datasets. Specifically,  
On the RPE Data \cite{nanni2016texture}, the OA of MambaMIC-T is comparable to FastViT-ma36 \cite{vasufastvit2023}, but our parameters and GMACs are 4.32 and 0.67, respectively, compared to FastViT-ma36’s 44.07 and 6.07. This represents a remarkable reduction of 90.2\% in parameters and 89.0\% in computation.  
On TissueMNIST \cite{medmnistv2}, MambaMIC-S improves accuracy by 6.54\% compared to ConvNext-Tiny \cite{liu2022convnet} while reducing parameters and computation by 83.1\% and 82.7\%, respectively.  
On MHIST \cite{Wei2021}, MambaMIC-T improves accuracy by 9.96\% over VMamba-T \cite{liu2024vmamba} while saving 79.7\% of parameters and 85.5\% of computational load.  
Notably, our model performs exceptionally well on large datasets (e.g., TissueMNIST \cite{medmnistv2}) and achieves optimal performance on small datasets (e.g., RPE Data \cite{nanni2016texture}). This highlights MambaMIC’s capability to efficiently handle microscopic image recognition tasks without requiring extensive data or computational resources. 

\subsection{Ablation Study}
In this section, we conduct ablation experiments on the RPE Data \cite{nanni2016texture} and MHIST \cite{Wei2021} to investigate the impact of individual components on the final performance. For a fair comparison, all ablation studies are performed under identical settings and training configurations.
\vspace{-2mm}
{\flushleft\textbf{Ablation experiments with different components}.}
To validate the effectiveness of the proposed components, we conducted a detailed ablation study in Table \ref{ablation1}. The LAEF significantly improves accuracy while further reducing model parameters and computational complexity. This is achieved through the channel clearing mechanism, which alleviates information redundancy and enhances the local perceptual ability of the VSSM, mitigating the local pixel forgetting issue. Additionally, the FMIAM achieves deep information fusion and key expert information relocation with fewer parameters, leading to a notable performance improvement.
\vspace{-2mm}
{\flushleft\textbf{Ablation study of the parallel VSSM mechanism}.}
In Table \ref{ablation2}, we further analyze the parallel VSSM mechanism, which reduces the parameter count by half while maintaining high performance. This is because the parallel mechanism not only effectively alleviates the computational burden caused by an excessive number of hidden states, but also promotes the interaction of channel context information.
\vspace{-2mm}
{\flushleft\textbf{Ablation study of LAEF}.}
To further validate the effectiveness of LAEF, we replaced it with a Linear layer (baseline), 3 \(\times\)3 convolution, Convolutional Gated Linear Unit (ConvGLU) \cite{shi2023transnext}, and Depth-wise Convolution Equipped Feed-forward Network (DFN) \cite{li2021localvit}. The quantitative results, shown in Table \ref{table:LAEF}, demonstrate that our LAEF achieves the best performance in both parameters and accuracy.

\vspace{-2mm}
{\flushleft\textbf{Ablation analysis of partial channel ratio}.}
In LAEF, we enhance local perception by retaining only a subset of channels through channel dropping and pruning mechanisms. The choice of partial channel rate \( r \) is therefore crucial, and we further analyze its selection. The experimental results, shown in Fig. \ref{rate}, reveal that when \( r \) is set too large or all channels are selected, no significant performance gain is observed. On the contrary, the redundancy of information increases the computational burden and introduces noise, which interferes with the accurate localization of key features. When \( r \) is set too small, valuable information is lost. Consequently, we explore the optimal ratio, and when \( r \) is set to 1/4 (the default setting for MambaMIC), it achieves the best trade-off between accuracy and speed.

\section{Conclusion}
In this paper, we explore the power of Mamba in MIC and reconsider its limitations. Specifically, we design a Local-Global dual-branch architecture, the MambaMIC Block. The local branch uses convolutions to enhance perception, while the global branch employs VSSM to capture global dependencies, incorporating LAEF to reduce channel redundancy and pixel forgetting. Additionally, FMIAM recalibrates features from both branches for multi-class fusion and key feature re-localization. Extensive experiments show that MambaMIC outperforms state-of-the-art methods, providing a new strong baseline for MIC.

\bibliographystyle{IEEEbib}
\bibliography{references-main}

\end{document}